\def\year{2021}\relax
\documentclass[letterpaper]{article} % DO NOT CHANGE THIS
\usepackage{aaai21}  % DO NOT CHANGE THIS
\usepackage{times}  % DO NOT CHANGE THIS
\usepackage{helvet} % DO NOT CHANGE THIS
\usepackage{courier}  % DO NOT CHANGE THIS
\usepackage[hyphens]{url}  % DO NOT CHANGE THIS
\usepackage{multirow}
\usepackage{amsmath}
\usepackage{bm}
\usepackage{graphicx} % DO NOT CHANGE THIS
\usepackage{natbib}  % DO NOT CHANGE THIS AND DO NOT ADD ANY OPTIONS TO IT
\usepackage{caption} % DO NOT CHANGE THIS AND DO NOT ADD ANY OPTIONS TO IT
\usepackage{booktabs}

\nocopyright
\title{Graph-based Multi-hop Reasoning for Long Text Generation}

\author{
Liang Zhao\thanks{\ \ Equal Contribution.}, \textsuperscript{\rm 1} 
Jingjing Xu\footnotemark[1], \textsuperscript{\rm 2} 
Junyang Lin, \textsuperscript{\rm 3} 
Yichang Zhang, \textsuperscript{\rm 3} 
Hongxia Yang, \textsuperscript{\rm 3} 
Xu Sun, \textsuperscript{\rm 1,2} \\
}
\affiliations{
\textsuperscript{\rm 1} Center for Data Science, Beijing Institute of Big Data Research, Peking University \\
\textsuperscript{\rm 2} MOE Key Lab of Computational Linguistics, School of EECS, Peking University \\
\textsuperscript{\rm 3} Alibaba Group \\
\{zhao1iang,jingjingxu,xusun\}@pku.edu.cn, \{junyang.ljy,yichang.zyc,yang.yhx\}@alibaba-inc.com
}

\begin{document}

\maketitle

\begin{abstract}
Long text generation is an important but challenging task. 
The main problem lies in learning sentence-level semantic dependencies which traditional generative models often suffer from. 
To address this problem, we propose a
% \textbf{Multi-hop Reasoning based Generative (MRG)}
\textbf{Multi-hop Reasoning Generation (MRG)} approach that
 incorporates multi-hop reasoning over a knowledge graph to learn semantic dependencies among sentences. MRG consists of two parts, a graph-based multi-hop reasoning module and a path-aware sentence realization module. The reasoning module is responsible for 
%  searching skeleton paths from a knowledge graph 
% searching paths from a knowledge graph as a skeleton 
searching skeleton paths from a knowledge graph
 to imitate the imagination process in the human writing for semantic transfer. Based on the inferred paths, the sentence realization module then generates a complete sentence. 
%  Owing to the lack of human annotation, 
%  we adopt a self-supervised approach to construct  pseudo parallel corpus 
%  to train the modules.
%  the lack of human-annotated training corpus, 
%  supervisory signals 
 Unlike previous black-box models, MRG explicitly infers the skeleton path, which provides explanatory views to understand how the proposed model works.  
We conduct experiments on three representative tasks, including story generation, review generation, and product description generation.
%  Both automatic and human evaluation demonstrate the effectiveness of the proposed model. The proposed model outperforms strong baselines significantly, especially in terms of informativeness, diversity and coherence.   
 Automatic and manual evaluation show that
our proposed method can generate more informative and coherent long text than
strong baselines, such as pre-trained models (e.g. GPT-2) and knowledge-enhanced models. 
% Our analyses demonstrate the inferred skeleton paths can efficiently build semantic dependencies among sentences and provide explanatory views to understand the generation process.
%Long text generation is an important but challenging task. The challenge lies within how to efficiently learn semantic dependencies among sentences. Here we introduce a \textbf{multi-hop Reasoning Generation (MRG)} approach that can  perform  multi-hop reasoning in knowledge graph to model the complex semantic relation between sentence to generate and the context and provide explanatory reasoning path for long text generation.
% The key idea is to model the reasoning process during writing like human being. 
%MRG can efficiently explore useful reasoning path for text generation with a self-supervised way, even there exists no  high-quality artificial knowledge graph.  With the help of the semantic relationship established by the reasoning path among sentences in knowledge graph, our approach can generate more coherent and informative long text. 

\end{abstract}

\section{Introduction}
% point out the importance , such as entertainment, e-commerce, and so on is an important but challenging task and is applicable in a number of fields. It

Long text generation is an important task and is applicable in a number of fields. It requires a model to generate a coherent and informative long text sequence which consists of several sentences based on free-format source text, such as a table, a topic sentence, or even some keywords~\cite{regina-et-al:table2text,nm:story_generation_2006,feng-et-al:2018topic}. 
% Long text generation techniques provide an automatic way to generate a large amount of high-quality contents, which can improve the efficiency of human writing. 
Recently, these techniques have been applied to many real-world tasks, like automatic story generation~\cite{nm:story_generation_2006}, review generation~\cite{reviewe-generation} as well as product description generation \cite{chen-et-al:product_description_generation}. %Since long text generation has great potential in application, it has become one of the focuses in both research and industry.
% and recent technology development in deep learning has made the generation of long text more possible than ever
% a simple definition
% Long text generation aims at generating coherent and cohesive long text with multiple sentences, such as story \cite{nm:story_generation_2006}, review \cite{edmund-et-al:wong2013autocomment}, and essay \cite{feng-et-al:2018topic}. 

% To be more specific, an automatic long text generation system is required to generate a meaningful coherent and informative long text based on several keywords or a topic sentence. 
% an automatic natural language generation system is required to generate a meaningful long  text based on the given information, such as keywords or a topic sentence. 
% of a story and the basic information of an item. 
% The sentences that it generates should be related to the input information and mutually connected on multiple aspects, including syntax, semantics and pragmatics. 

% are either a conditional language model \cite{gpt2} or a model with the sequence-to-sequence framework \cite{Seq2Seq,Transformer}. They 

\begin{figure}[t] 
\centering
\footnotesize
\includegraphics[width=1.0\linewidth]{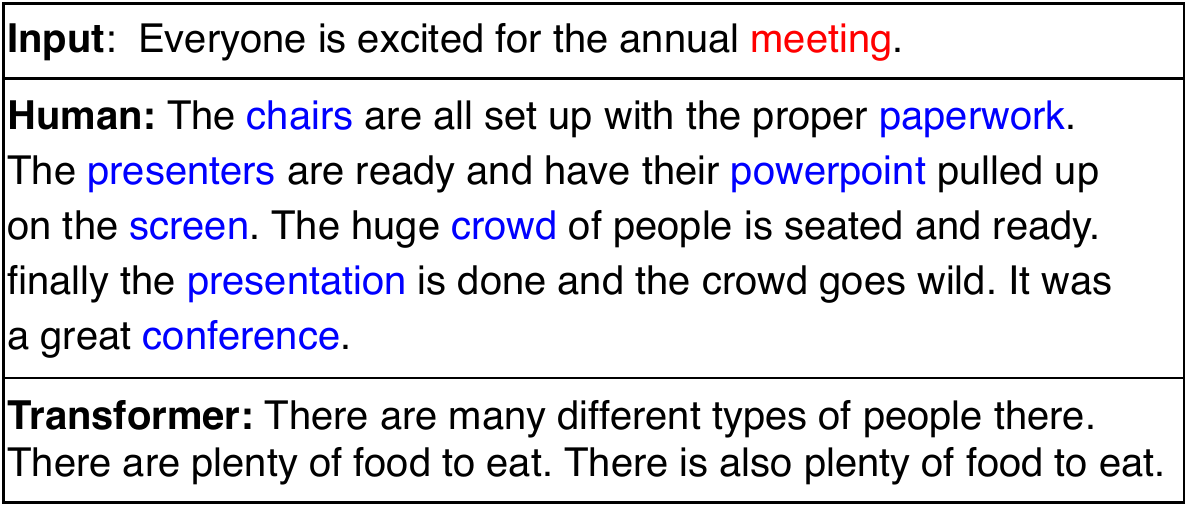}
\caption{The comparison between stories generated by a human and a widely-used generative model Transformer. As shown in the example, human can create diverse and coherent sentences with their imagination. For illustration, we highlight several skeleton words important for sentence-level semantic dependencies. In contrast, the machine-generated text is boring and repetitive. % Red word is concept in input through which humans imagine a lot of related concepts (blue words) for text generation. In contrast, machine-generated (transformer) text is boring and repetitive due to lack of imagination 
\label{fig:intro_example}}
\end{figure}

%Most of the current models have little power to model such dependencies.
Long text generation is more challenging than single sentence generation~\cite{langkilde-et-al:sentece_generation} in that it requires a model to learn the sentence-level semantic dependencies. Compared with word-level semantic dependencies in a single sentence,
% with short distance
sentence-level semantic dependencies are a kind of long-distance dependency that are harder to learn. 
Current end-to-end generative models~\citep{sutskever-et-al:seq2seq,vaswani-et-al:transformer} often adopt an attention mechanism to model the context dependencies, which fuses all word information together to generate the next word in a weighting way. 
Despite promising results in single sentence generation, these models suffer from complicated and sparse dependencies in long text generation with much longer contexts. Such dependencies make models hard to learn sufficient knowledge to infer semantic transfer among sentences.  Thus, these models tend to generate repetitive and boring sentences like a context-free language model, as shown in Figure~\ref{fig:intro_example}. In comparison, human can write diverse texts with coherent semantic transfer among sentences with their imagination. % about how new concepts can be transferred from  previously generated concepts, which current end-to-end generative models have not learned well.
Imagination is said to be a part of literacy process and it highly contributes to writing \citep{colello:colello2007imagination}.

\begin{figure*}[t]
\centering
\small
\footnotesize
\includegraphics[width=1.0\linewidth]{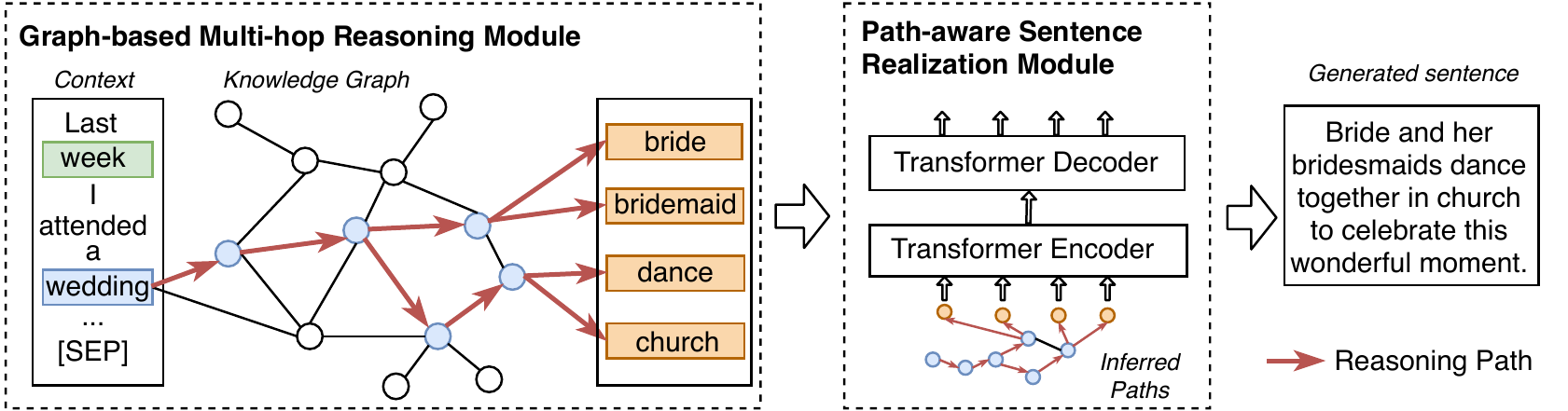}
% \includegraphics[\linewidth]{graph_trans.pdf}
% \vspace{-0.55in} 
\caption{An illustration about MRG in the inference stage. MRG 
% a sentence-level generative framework by 
iteratively generates a sentence based on previously generated contexts. During each iteration, the reasoning module infers skeleton paths starting from the concepts in the input context. Then this module iteratively searches the neighbor nodes of current path and predicts whether a neighbor node is target concept (ends of the inferred path), intermediate concept (intermediate nodes in the inferred path) or other concept (nodes not in the inferred paths). 
% The module continues searching only for intermediate concepts. 
The sentence realization module is responsible for generating a complete sentence based on the inferred paths. %Then the generated sentence is added to the context to generate the next sentence.
\label{fig:model}}
\end{figure*}

Motivated by human writing, we propose to endow text generation models with the imaginative power. To achieve this goal, we present a novel two-stage approach that imitates the semantic transfer in human writing. In the first stage, a graph-based reasoning module is responsible for inferring paths to select skeleton concepts based on the topic sentence and previously generated sentences via multi-hop reasoning on a knowledge graph. In the second stage, the path-aware sentence realization module approximates the target outputs based on the inferred paths. Owing to the lack of human-annotated training data, we adopt a self-supervised method to generate pseudo parallel corpus to train the two modules.
% Specifically, we treat paths in a knowledge graph from concepts in previously generated contexts to concepts in the target sentence as inferred paths. Then, we take the topic sentence and previously generated sentences as the input, and use the searched paths as the target to train the first module. We also take the searched paths as the input and the next sentence as the target to train the second module.
Experiment results on three long text generation tasks, including story generation,  review generation,  and  product  description  generation show that 
% We conduct experiments on three long text generation tasks, including story generation, review generation, and product description generation. Both automatic and human evaluation demonstrate 
 MRG consistently surpasses existing state-of-the-art generative baselines, such as large-scale pretrained models (e.g., GPT-2) and knowledge-enhanced text generation models, by significant margins, especially in terms of informativeness, diversity and coherence. 
 
 The main contributions are summarized as follows:

\begin{itemize}
\item We propose a novel multi-hop reasoning generation method to learn semantic dependencies among sentences. Unlike previous methods, our proposed model explicitly infers the skeleton path, which provides explanatory views to understand the process of generation. % Our proposed model explicitly learns the semantic dependency by conducting multi-hop reasoning in knowledge graph. 

% \item To the best of our knowledge, this is the first work to study how multi-hop reasoning helps text generation. We view this work as the first step towards learning semantic dependencies via multi-hop reasoning for long text generation.

\item Experimental results show that our approach outperforms state-of-the-art baselines especially in terms of informativeness, diversity and coherence.
% Our analyses demonstrate the efficiency of our method in long range semantic dependencies modeling.

% our works point a new direction for long text generation, i.e., imitating human writing process to model semantic connection among sentence  via multi-hop reasoning in knowledge graph.
% \item We build Self-constructed Graph for each task, so that MRG can perform complex reasoning even there exists no a ready-made knowledge graph.

\end{itemize}

\section{Proposal}
In this section, we introduce our proposed two-stage approach, \textbf{M}ulti-hop \textbf{R}easoning \textbf{G}eneration (\textbf{MRG}). We first give the formulation of long text generation and the related notations. Then, we provide an overview of our approach and introduce the two modules.

\subsection{Formulation and Notations}
% \medskip\noindent\textbf{Problem Formulation}\hspace{1em} 
% Suppose we have a parallel long text generation dataset 
% $\mathcal{P}$. 
Given a parallel long text generation dataset, each parallel sample consists of a source context $x$, and a corresponding target $\bm{y}$ containing $n$ sentences $\{y_1, y_2,\dots, y_n\}$. Formally, long text generation requires the generation system to generate an 
% grammatical, semantically coherent and pragmatically cohesive 
informative and coherent
long text $\bm{y}$ based on the source input $x$. %The training objective for the model is that the generation should approximate the target $y$. 
 
MRG adopts a sentence-level generative framework to learn semantic dependencies among sentences, which takes both the source context $x$ and the previous generation $\bm{y_{<j}}$ as the input when generating the $j$-th sentence $y_j$. Besides, a knowledge graph $\mathcal{G}$ is needed for MRG to conduct semantic transfer via multi-hop reasoning, and $\bm{v}=\{v_1,\dots,v_m\}$ is the set of  $m$ nodes on $\mathcal{G}$.

\subsection{Overview}
The architecture illustration of the proposed MRG model is shown in Figure~\ref{fig:model}. 
% MRG consists of two modules: the graph-based multi-hop reasoning module and path-aware sentence realization module. 
% We adopt a sentence-level text generation framework where the complete text is generated sentence by sentence.  
% MRG generate the complete text sentence by sentece.
%takes the previously generated context as input and
At each sentence-level generation,
the reasoning module first infers skeleton paths from a knowledge graph to imitate the imagination in the human writing for coherent sentence-level semantic transfer. 
Specifically, the inferred paths start from the concepts in the input sentences for initialization. The reasoning module iteratively adds new nodes into the inferred paths. At each reasoning step, the module first finds all neighbor nodes of current inferred paths as candidate nodes and predicts their labels. We define three kinds of labels, including \textbf{target concept} (ending of the inferred path), \textbf{intermediate concept} (middle node in the inferred path), and \textbf{other concept} (node not in the inferred path). For intermediate concepts, the module adds them into the inferred path and keeps the reasoning process. 
For target concepts, the module adds them to the inferred path and stops the reasoning process. For other concepts, the module just stops the reasoning process. We treat the inferred skeleton paths as pivots connecting semantics among sentences. Then, the realization module takes the inferred paths as inputs and decodes a complete  sentence. %The details of two modules and training algorithms are shown in the following.  

\subsection{Graph-based Multi-hop Reasoning}
The graph-based multi-hop reasoning module is responsible for learning the semantic dependencies among sentences via conceptual paths. 
This module explores on the knowledge graph to find a path that can connect concepts in the context with concepts in its subsequent sentence through intermediate concepts. 
Specifically, for a concept $v_1$ in a source sentence and $v_2$ in its subsequent sentence, their connected path is essential to semantic transfer.
% In our multi-hop reasoning method, we first extract concepts from both the source and target context.
% , which are labeled as \textbf{source concepts} and \textbf{target concepts}. 
% Specifically, for sentence-by-sentence long text generation, the source context is the previous generation, and the target of the next sentence to generate. 
MRG performs reasoning on a knowledge graph, which starts at the concepts in the source sentence and ends at the target concepts. The inferred concept chains form multiple conceptual reasoning paths.
% The concepts on the paths are labeled as \textbf{intermediate concepts}, and the others are labeled as \textbf{other concepts}. 
% With the training data and the labeled concepts, the reasoning module performs conditional sequence labeling. 
In the implementation, the reasoning module performs sequence labeling for nodes in a knowledge graph conditional on source input.
Concretely, at the $1$st hop, the input for the module is a sequence of the source context and their concepts on the graph $\bm{T^{(1)}}=\{x_1, \dots,x_{n},\mathtt{[SEP]},v^{(1)}_1,\dots,v^{(1)}_m\}$. The reasoning module learns to predict their labels of the concepts. 
When reaching an intermediate concept, MRG moves to the next hop and collects its neighbors.
% and adds the intermediate concepts and target concepts to the source context. 
The reasoning of neighbors continues and the module repeats the learning of sequence labeling until it meets concepts other than intermediate ones or the module reaches the max number of hop.
% already performed $k$ hops.  
In the inference stage, the reasoning module infers multiple paths and finds out possible target concepts and intermediate concepts. 
% More specifically, starting from any source concept, the module searches its neighbors and predict their labels. Unless it predicts an intermediate concept, the search at that path ends. For a prediction of intermediate concept, the search continues and repeats the aforementioned process. 
We adopt the powerful pretrained model BERT~\cite{devlin-et-al:bert} as the backbone. Below we provide more details of the module. 

\medskip\noindent\textbf{BERT-based Labeller}\hspace{1em}
The module is a sequence labeling model with BERT as the encoder. 
The goal of BERT encoder is to provide a representation for conditional sequence labeling. At the $k$-hop reasoning step, it takes $\bm{T^{(k)}}=\{x_1, \dots,x_{n},\mathtt{[SEP]},v^{(k)}_1,\dots,v^{(k)}_{l}\}$ as the input, where $l$ is the number of neighbors of the previous intermediate nodes, and generates hidden representations:
\begin{equation}
\bm{h^{(i)}}= BERT(\bm{T^{(i)}}),
\end{equation}
where $\bm{h^{(i)}}$ is a vector sequence. Each vector is the representation of the word at the corresponding position. 
For  simplicity, we omit the details of BERT, and refer the readers to the original work~\cite{devlin-et-al:bert}.

% \medskip\noindent\textbf{Labeller.}\hspace{1em}
On top of the BERT encoder, we add a classifier layer for labeling based on Multi-Layer Perceptron (MLP). The input of the classifier is the  representation in the last layer of the BERT encoder $\bm{h^{(i)}}$. The following computation demonstrates how to predict the label $l_r$ at position $r$:
% The output representation of each position is then sent to a classifier to predict label of each words. Comply with the recommendations in the original Bert paper, we adopt a simple MLP to compute the probability distribution of the label $p_i(l_i|T_i)$:
\begin{equation}
p(l_r|\bm{h^{(i_r)}}) = softmax(\bm{W_c}\bm{h^{(i_r)}}+b_c),
\end{equation}
where $\bm{W_c}$ and $b_c$ are the weight matrix and bias term. We mask  the losses of  context words when doing sequence labeling.

% \medskip\noindent\textbf{Data Collection for Module Two}\hspace{1em}
% After the process of multi-hop reasoning, the target concepts are added to the target concept set $W^T_{set}$, and intermediate concepts are added to the intermediate concept set $W^I_{set}$. The two sets of words are significant inputs for the module, which is illustrated in the following.
% \subsubsection{Construct Supervisory Signals}

\subsection{Path-aware Sentence Realization}
The path-aware sentence realization module aims at generating a sentence that approximates the target based on the inferred paths. 
% the module for sentence realization is based on the sequence-to-sequence framework, 
We adopt the architecture of Transformer~\cite{vaswani-et-al:transformer} for sentence realization.
% due to its supreme performances in a series of generation tasks. 
In the following, we illustrate the details of data preparation and module architecture. 
% This module follows the traditional sequence-to-sequence framework, and we adopt Transformer  as the sentence realization module. 

\medskip\noindent\textbf{Data Preparation}\hspace{1em}As shown in Figure~\ref{fig:model}, we extract the inferred paths from the knowledge graph $G$ to build a sub-graph $G'_{x,y<t}$ for sentence realization. Specifically, for each concept $\hat{n}^{(i)}$ in inferred target concepts set $\bm{\hat{n}}$, without changing the topology of the nodes that directly connect to $\hat{n}^{(i)}$ on $G$, we also connect the intermediate nodes in the reasoning path which can point to $\hat{n}^{(i)}$ with $\hat{n}^{(i)}$. In this sub-graph all the intermediate concepts directly connect with its corresponding target concepts.
% As shown in Figure \ref{fig:model}, we build a sub-graph $G'_{x,y<t}$ as 
% the input for sentence realization based on the concepts in source text $x$ and  the previous target text $\bm{y_{<t}}$ (it refers to the previous generation during the inference stage), the target word sets $W^T_{set}$ and intermediate words set $W^I_{set}$ and the given knowledge Graph $G$. Specifically, We take concepts in $W^T_{set}$ as main concepts, denoting as $\hat{n}$. For each concept $\hat{n}^i$ in $\hat{n}$, we connect the concepts in the inferred path which point to $\hat{n}^i$ and the concepts in $W^I_{set}$, $x$ and $\bm{y_{<t}}$ which  directly connect with the $\hat{n}^i$ in $G$ with $\hat{n}^i$ to build the sub-graph $G'_{x,y<t}$.

% firstly we connect the concepts in the inferred path which point to $\hat{n}^i$ and the concepts in $W^I_{set}$, $x$ and $\bm{y_{<t}}$ which  directly connect with the $\hat{n}^i$ in $G$ with $\hat{n}^i$ to build the sub-graph $G'_{x,y<t}$.

\medskip\noindent\textbf{Transformer}\hspace{1em} It builds upon on a encoder-decoder framework.
% As it is limited in generating long texts, we use Transformer to generate the text sentence by sentence. 
For each node in target node set $\bm{\hat{n}}$ of the input graph $G'_{x,y<t}$, we concatenate its directly linked nodes as its features. Then we use the sum of target node embedding and its feature node embeddings as node representation. The node representation of $\bm{\hat{n}}$ is denoted as $\bm{\hat{e}}$. On the transformer encoder, we find modeling the order of paths is more complicated but cannot improve the final results, so we remove the positional encoding to eliminate the effects of reasoning order.
The Transformer encoder (TrmEnc) encodes $\bm{\hat{e}}$ to a list of representations $\bm{\hat{h}}$: 
% For the target words in $G^{`}$, we add all connected intermediate words embedding to corresponding target word embedding which can be denoted as $V^o=\{v^o_0, v^o_1,\dots, v^o_m\}$. Then we use a special token ``[SEP]'' to separate  source sentence and target words. The input can be formulated as $q=\{x,\bm{y_{<t}},[SEP],V^o\}$
\begin{equation}
\bm{\hat{h}} = TrmEnc(\bm{\hat{e}}). \\
\end{equation}
At the $i-{th}$ decoding time step, the Transformer decoder (TrmDec)  generates a probability distribution for word prediction by attending to both $\bm{\hat{h}}$ and previous decoder outputs:
\begin{equation}
\begin{split}
p(y_{i,t}|\bm{y_{<i,t}}, \bm{\hat{h}}) &= TrmDec(\bm{y_{<i,t}}, \bm{\hat{h}}). \\
% p(y_t|y_{<t}, h^j) &= softmax(W_pv_t + b_p)
\end{split}
\end{equation}
Both the encoder and decoder consist of multiple layers, each with a multi-head self-attention sub-layer and a point-wise feed-forward neural network (FFN) sub-layer, and each of the layers of the decoder has an additional encoder-decoder attention sub-layer.
% \subsection{Training and Inference}
% In the training stage, 
% % the objective functions of 
% the two modules are optimized using cross entropy loss.
% In the inference stage, the two modules work together following a two-stage framework. In the first stage, the reasoning module infers several conceptual paths and emits the paths with the realization module . In the second stage,  the realization module decodes a sentence based on the inferred paths.

\medskip\noindent\textbf{Pseudo Parallel Data Construction}\hspace{1em}Owing to the lack of human-annotated training corpus, we adopt a self-supervised approach to construct peseudo parallel data to train our two modules. 
% We extract data from the training set and build a dataset. 
% Specifically, similar to the data for next sentence prediction in Devlin et al.~\citet{devlin-et-al:bert}, we extract sentence pairs from a text with multiple sentences.
% Specifically, we treat paths in a knowledge graph from concepts in previously generated contexts to concepts in the target sentence as inferred paths. Then, we take the topic sentence and previously generated sentences as the input, and use the searched paths as the target to train the first module. We also take the searched paths as the input and the next sentence as the target to train the second module.
% We extract sentence pairs from a text with multiple sentences in training data. 
% In fact, the former one in the pair can be a longer paragraph with multiple sentences, and the latter one is the subsequent sentence. 
% For training the first reasoning modules, we define the co-occurence entity when starting from the source sentence go through several intermediate nodes in knowledge graph that can meet entity in target sentence, then we call the nodes in target sentence as target nodes, and the other nodes in this path as intermediate nodes. 
% Algorithm 1 illustrates the process of automatically annotating the reasoning sequence. 
% Given a long text, We first cut the long text into the form of sentence pairs, of each pair consist of directly connected sentences, we call them  source sentence and target sentence.
Given a long text, we first cut the long text into the form of sentence pairs, of each pair consist of two consecutive sentences.
For each sentence pair, we search the concepts that exist in the knowledge graph. If a concept $v_1$ in the former sentence can reach a concept $v_2$ in the latter sentence via $k+1$ hops, the path with $k+1$ nodes are labeled. To be specific, $v_2$ is labeled as target concepts, $v_1$ and $k$ intermediate nodes are labeled as intermediate concepts. We set a constraint for the  max number of hops and neighbors of a concept to avoid over-exploration. Then, we collect  pairs with at least one labeled path to form a dataset for sequence labeling. Besides, we collect the pairs of paths with its subsequent sentence to form a dataset for sentence realization.

\subsection{Self-constructed Graph for MRG}
Although there are several widely-used artificial knowledge graphs, such as ConceptNet~\cite{liu-singh:concept_net}. However, for some specific fields, 
% we are still lacking in high-quality knowledge graphs. 
we still lack in an artificial knowledge graph. 
To make MRG applicable in such scenarios,  we propose to build a graph based on the training data. Inspired by~\citet{textgcnPMI,pmigraph}, we build a graph via PMI (Point-wise Mutual Information)~\cite{andrew-et-al:pmi3}, a popular measure for word associations. The behind motivation is that words in the same text are more likely semantically connected, so we can connect those words with edges to build a graph called as Self-constructed Graph. Specifically, we compute the PMI between any two words in the training vocabulary except for stop words with:
% To tackle the problem, we propose a method to build a knowledge graph based on the training data. Specifically, we build a knowledge graph via PMI (Point-wise Mutual Information)~\cite{andrew-et-al:pmi3,su-et-al:pmi2}. The behind motivation is that words in the same text more likely share the same topic, so we can connect those words with the same topic with edges and build a knowledge graph, which is called as Self-constructed Graph. Specifically, we compute the PMI between any two words in a vocabulary except for stop words with:
\begin{equation}
\begin{split}
PMI(w_i;w_j) = log\frac{p(w_i,w_j)}{p(w_i)p(w_j)} 
% p(y_t|y_{<t}, h^j) &= softmax(W_pv_t + b_p)
\end{split}
\end{equation}
where $p(w_i, w_j)$ refers to the probability of the two words $w_i$ and $w_j$ in the same text. $p(w_i)$ and $p(w_j)$ are the probabilities of words in the entire training data. For each words, we connect those words with top-$k$ PMI values.

\section{Experiments}
% We evaluate our method on three long text generation tasks, including story generation, review generation and product description generation.
% In the following, we introduce the details of our experiments.

\subsection{Dataset}
Here we describe the used datasets and  Table~\ref{tab:dataset} demonstrates the detailed statistics.

\medskip\noindent\textbf{Story Generation}\hspace{1em}This dataset comes from a visual storytelling dataset~\cite{huang-et-al:visualstory} which consists of pairs of an image sequence and a human-written story. 
Following ~\citet{xu-et-al:skeleton_jingjingxu_emnlp8}, we only use the text data for our experiments.
For each story, we extract the first topic sentence as the source input and the rest of the story as the target. 

\medskip\noindent\textbf{Review Generation}\hspace{1em}This dataset is provided by Yelp
Challlenge\footnote{https://www.yelp.com/dataset/challenge}.
% contains user reviews on Yelp.
Similarly, for each review, the first sentence is the source input and the rest is the target.

\medskip\noindent\textbf{Description Generation}\hspace{1em}This dataset is crawled from Taobao\footnote{www.taobao.com}, 
a Chinese e-commerce platform. It consists of pairs of product titles and the corresponding product descriptions, which are composed by professional content producers on the website. We discarded pairs with a description less than 20 tokens or longer than 150 tokens to remove noisy samples. We will release this dataset if this paper is publised.%~\footnote{.}

% The dataset statistics are shown in Table 3. For story generation, we use the standard split. For review generation and description generation, we randomly sample 5K, 5K as test and development set, the rest is  training set.

% \subsection{Evaluation Metrics}
% We evaluate the diversity and novel of each generated text by automatic evaluation. 
\subsection{Baselines}
% Here we implement several widely-used long text generation methods, and some pre-trained model as our baseline models. The details
% We compare our models with three types of baseline models: widely-used text generation models (Seq2Seq, Transformer), two stage skeleton-based method (SkeletonSeq) and the large scale pre-trained model (GPT2). 
We compare our models with the following widely-used text generation approaches.
\begin{itemize}
\item \textbf{Seq2Seq}~\cite{sutskever-et-al:seq2seq} with attention mechanism~\cite{bahdanau-et-al:attention} is a widely-used model for text generation.
\item \textbf{Transformer}~\cite{vaswani-et-al:transformer} is the state-of-the-art architecture for 
a series of tasks of nature language generation. 
% \item \textbf{Seq2Seq-K} is similar to Seq2Seq model, we concatenate knowledge entity that directly connected in knowledge graph with word in source sentence, we also enhance copy mechanism to this model called \textbf{Seq2Seq-KC}. 
% \item \textbf{Transformer-C} is similar to Transformer. We concatenate the source input with the concepts directly connected to words of the source in the knowledge graph. Furthermore, we apply Copy mechanism \cite{copy} to enhance the capabilities of the model. Specifically, \textbf{Transformer-CC} denotes the model trained with the concepts in ConceptNet, and  \textbf{Transformer-CI} denotes the model trained with the concepts in Self-constructed Graph.
% \item \textbf{GPT} is a large pre-trained model 
\item \textbf{Skeleton2Seq}~\cite{xu-et-al:skeleton_jingjingxu_emnlp8} is a two-stage generative method, which first generates a skeleton and then expands it to a complete sentence. This method also uses the sentence-level recurrent generation framework.
\item \textbf{Plan\&Write}~\cite{plan-and-write} is a two-stage generative method, which first generates a sequence of keywords as a storyline and then expands it to a complete story.

\item \textbf{GPT-2}~\cite{radford-et-al:gpt2} is a large Transformer-based language model with 12 layers and  110M parameters, which has reached state-of-the-art performances on many language generation tasks. We fine-tune GPT-2 on training data for better performance, and name the model \textbf{GPT-2-FT}.

\item \textbf{GPT-2-KE}~\cite{gpt2-ke} is a knowledge-enhanced pre-training model, which incorporates commonsense knowledge  into GPT-2  and adopts a multi-task learning method during fine-turning  to improve coherence.
\end{itemize}
% For a fair comparison, all the encoder-decoder based baseline use the same sentence-level recurrent generation framework and for GPT-2-FT we set the maximum  length of generation to 100. 
\subsection{Implementation Details}
Based on our preliminary experiments, 
% the vocabulary size for the source and target are 15K and 16K respectively. 
 the size of vocabulary is limited to 16,000. For each concept we explore up to 20 neighbors in the knowledge graph without going through loops. $k$ in Self-constructed graph is set to 20. We set the maximum reasoning number to 3. Both the modules are optimized with the Adam  optimizer~\cite{kingma-ba:adam}.
We fine-tune the BERT-based labeller with  the learning rate of 2e-5 and the batch size of 64 for 3 epochs. 
We implement the sentence realization module with a two-layer encoder and a two-layer decoder. We train this module with the learning rate of 0.001 and the batch size of 64 for 50 epochs.
We set the size of hidden dimension and embedding dimension to 128.
The number of heads in multi-head attention is 8.
The dropout rate is tuned in range of $\{0.1, 0.2, 0.3\}$.
% and we clip the gradients when the norm exceeds 5.
% We choose the last checkpoint for evaluation.  
% For a fair comparison, 
We adopt greedy sampling for all the models during decoding. 
We also use the same hyper-parameters and settings for Seq2Seq and Transformer.
% For baseline the max generated length is set to 100, and our model generate five sentences.
For GPT2-FT, we use the hyper-parameters provided by Huggingface~\cite{wolf-et-al:huggingface}. For other baselines, we use the released code and adopt the default settings for implementation.
% The learning rate is 5e-5 and the batch size is 32. The fine-tuning lasts 3 epochs\footnote{The default fine-tuning epoch is 1, but we find that fine-tuning GPT-2 for 3 epochs achieves a better performance.}.

%%%%%%%%%%%%%%%%%%%data%%%%%%%%%%%%%%%%%%%%%%%%%%%%%
\begin{table}[t!]
  % \begin{tabular}{c | c | c | c | c | c | }
  \centering
  \small
  \scalebox{0.95}{\begin{tabular}{lccccc}
  \toprule  %添加表格头部粗线
  \textbf{Dataset} & \textbf{Avg. \#w} & \textbf{Avg. \#s} & \textbf{Train} & \textbf{Dev} & \textbf{Test}   \\
  \midrule  %添加表格中横线
   Story  & 44.8 & 4.1 & 40K & 5K & 5K \\
   Description & 55.1 & 2.9 & 342K & 5K & 5K  \\
   Review  & 64.8 & 4.8 & 1,086K & 5K & 5K  \\
  \bottomrule %添加表格底部粗线
  \end{tabular}}
  \caption{Dataset statistics. ``Avg. \#w'' and ``Avg. \#s'' are the average amounts of word and sentence per text, respectively. ``Train'', ``Dev'', and ``Test'' are the sizes of the training set, development set, and test set.  }
  \label{tab:dataset}
  \end{table}
%%%%%%%%%%%%%%%%%%%data%%%%%%%%%%%%%%%%%%%%%%%%%%%%%

\begin{table}[t!]
  % \begin{tabular}{c | c | c | c | c | c | }
  \centering
  \small
  \scalebox{0.95}{\begin{tabular}{lccc}
  \toprule  %添加表格头部粗线
  \textbf{Knowledge Graph}  & \textbf{\# Nodes} & \textbf{\# Triples} &\textbf{Language}\\
  \midrule  %添加表格中横线
    Story-S   & 16,000 & 318,823 & English \\
    Review-S  & 16,000 & 320,000 & English\\
    Description-S  & 16,000 & 320,000 & Chinese \\
    ConceptNet & 21,471 & 149,809 & English \\
  \bottomrule %添加表格底部粗线
  \end{tabular}}
  \caption{Knowledge graph statistics. ``Story-S'', ``Review-S'' and ``Description-S'' are the self-constructed graphs for story, review and description generation datasets, respectively. ``\textbf{\# Nodes}''  and ``\textbf{\# Triples}'' are the total number of nodes and triples. }
  \label{tab:knowledge_graph}
  \end{table}

% \begin{table}[t!]
% 	\centering
% 	\small
% % \scriptsize
% 	%\setlength{\tabcolsep}{3pt}
% 	\begin{tabular}{@{}lcccc@{}}
% 		\toprule
% 		\textbf{Story Generation}&
%         \textbf{Token}& 
% 		\textbf{Dist-1}&
%         \textbf{Dist-2}&
%         \textbf{Dist-3} \\
%         \midrule
%           Seq2Seq   &  145K & 1.3K & 4.6K & 10K   \\
%           Transformer & 138K & 1.3K & 5.6K & 13K \\
%           Transformer-K & 143K & 0.8K & 2.8K & 6.5K   \\
%           Transformer-CC & 160K & 1.0K & 3.5K & 7.9K    \\
%           Skeleton & * & * & * & *  \\
%           GPT-2  & 214K & 2.1K & 7.6K & 16.7K  \\
          
%           \midrule
%           MRG-C & 207K & 2.0K & 11K & 27.4K  \\
%           MRG-S & 211K & 3.0K & 20K & 54.1K   \\
%     \midrule  \midrule
%     Web-Data Generation 
%     \bottomrule
% 	\end{tabular}
% 	\caption{Results of Diversity. Token denotes the number of generated words. Dist-n denotes the number of distinct n-gram in the generated text}
	
%     \label{table:human_evaluation}
%     % \vspace{-0.2in}
% \end{table} 

%%%%%%%%%%%%%%%%%%%%%%Automatic Evaluation%%%%%%%%%%%%%%%%%%%%%%%
\begin{table}[t]
  \centering
  \small
  \setlength{\tabcolsep}{1.8pt}
  \scalebox{0.85}{\begin{tabular}{lrrrrrrrr}
    \toprule
		\textbf{Story}&
          \textbf{Token$\uparrow$} & 
          \textbf{Dist-1$\uparrow$} &
          \textbf{Dist-2$\uparrow$} & 
          \textbf{Dist-3$\uparrow$} &
          \textbf{Uni-J$\downarrow$} &
          \textbf{Bi-J$\downarrow$} &
          \textbf{Tri-J$\downarrow$} &
          \textbf{BLEU$\uparrow$} \\
        \midrule
          Seq2Seq   &  158K & 1.3K & 4.6K & 10.2K & 71.6 & 50.1 & 38.6 & 2.36  \\
          Transformer & 138K & 1.3K & 5.6K & 13.3K & 67.7 & 46.4 & 35.7& 2.34\\
        %   Transformer-K & 143K & 0.8K & 2.8K & 6.5K   \\
        %   Transformer-CC & 177K & 1.7K & 7.5K & 17.7K & 61.1 & 37.0 & 26.5    \\
          Skeleton2seq & 160K & 2.0K & 9.4K & 21.1K  & 52.6 & 30.0 & 21.6 & 2.95 \\
          Plan\&Write  & 159K & 1.4K & 8.6K & 20.3K & 38.9 & 19.3 & 13.2 & \textbf{3.35} \\
          GPT-2-FT  & 216K & 2.1K & 7.6K & 16.8K & 34.6 & 21.4 & 15.7 & 2.73 \\
          GPT-2-KE  & 158K & 2.4K & 10.7K & 26.0K & 35.4 & 21.7 & 16.0 & 3.18 \\
          \midrule
          MRG-C (ours) & \textbf{284K} & 1.7K & 10.9K & 31.9K & 35.7 & 18.1 & 12.9 & 3.09 \\
          MRG-S (ours) & 259K & \textbf{3.3K} & \textbf{26.8K} & \textbf{75.1K} & \textbf{28.9} & \textbf{12.1} & \textbf{8.1} & 3.23  \\
         \midrule  \midrule
          \textbf{Review} &
          \textbf{Token$\uparrow$} & 
          \textbf{Dist-1$\uparrow$} &
          \textbf{Dist-2$\uparrow$} & 
          \textbf{Dist-3$\uparrow$} &
          \textbf{Uni-J$\downarrow$}&
          \textbf{Bi-J$\downarrow$} &
          \textbf{Tri-J$\downarrow$} &
          \textbf{BLEU$\uparrow$}\\
          \midrule
          Seq2Seq   &  201K & 0.3K & 1.1K & 2.4K &  60.2 & 31.5 & 24.6 & 1.28 \\
          Transformer & 210K & 0.4K & 1.7K & 3.7K & 62.1 & 30.5 & 22.3 & 1.15\\
        %   Transformer-K & 115K & 0.3K & 1.0K & 1.9K   \\ 
        %   Transformer-CC & 282K & 0.6K & 2.6K & 5.7K & 56.5 & 26.8 & 20.1     \\
          Skeleton2seq & 156K & 0.7K &  2.1K & 4.2K & 61.8 & 41.0 & 31.7 & 1.65\\
          Plan\&Write & 125K & 0.9K & 4.1K & 8.4K & 40.7 & 14.4 & 10.3 & 1.72\\
          GPT-2-FT  & 171K & 1.9K & 6.7K & 14.0K & 43.8 & 18.4 & 14.4  & 1.47\\
          GPT-2-KE  & 159K & \textbf{2.3K} & 8.7K & 18.0K & 44.0 & 28.9 & 20.7 & \textbf{2.24} \\ 
          \midrule
          MRG-C (ours) & 328K & 1.4K & \textbf{9.6K} & \textbf{28.1K}  & \textbf{32.9}  & \textbf{14.2} & \textbf{9.8} & 1.93 \\
          MRG-S (ours) & \textbf{375K} & 1.6K & 6.5K & 14.9K & 42.9 & 22.1 & 15.0 & 2.06  \\
          \midrule 
          \midrule 
          \textbf{Description} &
          \textbf{Token$\uparrow$} & 
          \textbf{Dist-1$\uparrow$} &
          \textbf{Dist-2$\uparrow$} & 
          \textbf{Dist-3$\uparrow$} &
          \textbf{Uni-J$\downarrow$}&
          \textbf{Bi-J$\downarrow$}&
          \textbf{Tri-J$\downarrow$}&
          \textbf{BLEU$\uparrow$}\\
          \midrule
          Seq2Seq   &  259K & 0.6K & 2.1K & 4.0K  &  42.2 & 21.5 & 15.0 & 1.64    \\
          Transformer & 224K & 1.3K & 4.5K & 8.9K & 37.5 & 18.7 & 13.2 & 1.77 \\
        %   Transformer-K & 245K & 1.2K & 3.8K & 7.2K   \\
        %   Transformer-CI & 237K & 1.5K & 5.5K & 10.9K & 38.7 & 19.7 & 13.8    \\
          \midrule
          MRG-S (ours) & \textbf{311K} & \textbf{3.0K} & \textbf{17.0K} & \textbf{37.2K}  &\textbf{31.8} & \textbf{16.5} & \textbf{11.6} & \textbf{2.44} \\ 
          \bottomrule %&\textbf{73.48}  
  \end{tabular}}
  \caption{Results of automatic evaluation. We evaluate the diversity and novelty using Dist-n and n-Jaccard respectively. Higher Dist-n represents higher diversity, and lower n-Jaccard represents higher novelty. \textbf{Token} denotes the amount of generated words. \textbf{Dist-n} denotes the number of distinct n-grams in the generated text. \textbf{Uni-J}, \textbf{Bi-J} and \textbf{Tri-J} represent the Jaccard similarity scores based on unigram, bigram and trigram. }
\label{table:automatic}
\end{table} 

\begin{table}[t]
\small
\centering
  \scalebox{0.95}{\begin{tabular}{|l|r|r|r|r|c|c|}
    \hline
    
    \textbf{Story}& \textbf{Fluen.}  & \textbf{Infor.} & \textbf{Coher.} & \textbf{G}& \textbf{$p$-value} \\ 
     
    \hline
     Seq2Seq & 58.3 &  42.7 & 41.7 & 47.6 &\multirow{2}{*}{*} \\
      \textbf{MRG-C}  &  \textbf{141.7} & \textbf{157.3}  & \textbf{158.3} & \textbf{152.4} &\\
      %\textbf{improvements} & & & & \\
  \hline
     Transformer &63.0  & 50.0  & 46.7 & 53.2 &\multirow{2}{*}{*} \\
      \textbf{MRG-C}  & \textbf{137.0}  &  \textbf{150.0} & \textbf{153.3}  & \textbf{146.8} &\\
      %\textbf{improvements} & & & & \\
       
      \hline
    
     Skeleton2Seq & 84.0 & 72.0  &  81.0 & 79.0&\multirow{2}{*}{*} \\
      \textbf{MRG-C}  & \textbf{116.0}  & \textbf{128.0} & \textbf{119.0} & \textbf{121.0}&\\
      %\textbf{improvements} & & & & \\
  \hline
       Plan\&Write & 77.0 & 81.3  &  87.3 & 81.9&\multirow{2}{*}{*} \\
      \textbf{MRG-C}  & \textbf{123.0}  & \textbf{118.7} & \textbf{112.7} & \textbf{118.1}&\\
      %\textbf{improvements} & & & & \\
  \hline
      GPT-2-FT & \textbf{108.3} &  72.7 &  84.0 & 91.7 &\multirow{2}{*}{**} \\
      \textbf{MRG-C}  &  81.7 & \textbf{127.3} & \textbf{116.0} & \textbf{108.3}&\\
      %\textbf{improvements} & & & & \\
  \hline 
      GPT-2-KE & \textbf{109.3} &  83.0 &  95.3 & 95.9 &\multirow{2}{*}{-} \\
      \textbf{MRG-C}  &  90.7 & \textbf{117.0} & \textbf{104.7} & \textbf{104.1}&\\
      %\textbf{improvements} & & & & \\
  \hline 
      \textbf{MRG-S} & 92.3 & \textbf{135.7}  & \textbf{117.0}  & \textbf{115.0} &\multirow{2}{*}{**} \\
      \textbf{MRG-C}  & \textbf{107.7}  & 64.3 & 83.0 & 85.0 &\\
  \hline
  \hline 
  \textbf{Review} & \textbf{Fluen.}  & \textbf{Infor.} & \textbf{Coher.}& \textbf{G}& \textbf{$p$-value} \\ 
    \hline
     Seq2Seq & \textbf{100.3} & 50.7  & 81.7 & 77.6 &\multirow{2}{*}{*} \\
      \textbf{MRG-C}  &  99.7 & \textbf{149.3}  & \textbf{118.3} & \textbf{122.4} &\\
      %\textbf{improvements} & & & & \\
  \hline
     Transformer & 91.7 &  50.3 & 78.3 & 73.4 &\multirow{2}{*}{*} \\
      \textbf{MRG-C}  & \textbf{108.3}  & \textbf{149.7}  &  \textbf{121.7} & \textbf{126.6} &\\
      %\textbf{improvements} & & & & \\
       
      \hline
    
     Skeleton2Seq & 95.7 &  72.0 & 75.0  & 80.9 &\multirow{2}{*}{*} \\
      \textbf{MRG-C}  & \textbf{104.3}  & \textbf{128.0} & \textbf{125.0} & \textbf{119.1}&\\
      %\textbf{improvements} & & & & \\
  \hline
      Plan\&Write  & 89.0 &  82.3 & 84.7 & 85.3 &\multirow{2}{*}{**} \\
      \textbf{MRG-C}  & \textbf{111.0}  & \textbf{117.7} & \textbf{115.3} & \textbf{114.7}&\\
      %\textbf{improvements} & & & & \\
  \hline
      GPT-2-FT & \textbf{ 105.3} & 82.3  & 91.0  & 92.9 &\multirow{2}{*}{**} \\
      \textbf{MRG-C}  & 94.7  & \textbf{117.7} & \textbf{109.0} & \textbf{107.1}&\\
  \hline
       GPT-2-KE & \textbf{ 105.0} & 85.0  & 87.3  & 92.4 &\multirow{2}{*}{**} \\
      \textbf{MRG-C}  & 95.0  & \textbf{107.0} & \textbf{112.7} & \textbf{107.6}&\\
  \hline
      \textbf{MRG-S} & 91.7 & 75.3  & 85.7  & 84.2 &\multirow{2}{*}{**} \\
      \textbf{MRG-C}  & \textbf{108.3}  & \textbf{124.7} & \textbf{114.3} & \textbf{115.8}&\\
      %\textbf{improvements} & & & & \\
  \hline 
  \hline 
  \textbf{Description} & \textbf{Fluen.}  & \textbf{Infor.} & \textbf{Coher.}& \textbf{G}& \textbf{$p$-value} \\ 
    \hline
     Seq2Seq & \textbf{100.3} &  69.7 & 87.3 & 85.8 &\multirow{2}{*}{**} \\
      \textbf{MRG-S}  & 99.7  &  \textbf{130.3} & \textbf{112.7} & \textbf{114.2} &\\
      %\textbf{improvements} & & & & \\
  \hline
     Transformer & \textbf{106.0} & 68.7  & 86.3 & 87.0 & \multirow{2}{*}{**}\\
      \textbf{MRG-S}  & 94.0  &  \textbf{131.3} &  \textbf{113.7} & \textbf{113.0} &\\
  \hline     
  \end{tabular}}
    \caption{Results of human evaluation. \textbf{Fluen.}, \textbf{Infor.} and \textbf{Coher.} denote fluency, informativeness and  coherence respectively.~\textbf{G} is the geometric mean of the three metrics . ``*'' means $p<0.01$ and ``**'' means $p<0.05$ in t-test. For each metric, the averaged Cohen's coefficient is in range of 0.3 to 0.6, which ensures inter-annotator agreements.} \label{tab:human evaluation}

\end{table}

Our method has two variants, \textbf{MRG-S} and \textbf{MRG-C}, which refer to MRG that performs reasoning on  Self-constructed Graph and ConceptNet~\cite{liu-singh:concept_net} respectively.
The statistics of the knowledge graphs are shown in Table~\ref{tab:knowledge_graph}.
% 备用 如果不加自建图在story 和 review上的结果的话就用这个
% For story and review generation, the proposed MRG conducts reasoning on ConceptNet~\cite{liu-singh:concept_net}, an external knowledge graph, which contains a large number of multilingual knowledge, and we only use the triples of the English version. The knowledge graph we used consists of 21,471 nodes and 149,809 triples in total. For product description generation, MRG conducts reasoning on Self-constructed Graph built on corresponding training dataset.

\subsection{Evaluation Metrics}
In this paper, we evaluate the quality of the generated texts by automatic and human evaluation. For automatic evaluation, because of the lack of reliable evaluation metrics in terms of coherence, we only evaluate the diversity and novelty of the generated texts and similarity between the generated texts with human texts. For human evaluation, we evaluate their coherence, informativeness and fluency.

% \subsubsection{Automatic Evaluation}
% \medskip\noindent\textbf{Metrics}\hspace{1em} 
% In long text generating task, not the more tokens is generated the better, e.i., generating boring and repetitive text is not wanted, which requires the model has the capacity to generated diversified text. 
% Besides, if a model is just memorize or copy sentences from the training data, having not the capability to learn the inner patterns of human writing is also not wanted, so generating novel text is also crucial. 
\medskip\noindent\textbf{Automatic evaluation}\hspace{1em}Long text generation expects the generated texts to be diversified and novel. 
%We evaluate the model performance with the metrics of diversity and novelty. 
\textbf{Diversity} is measured by the number of distinct n-grams in the generated texts. 
\textbf{Novelty} is measured by Jaccard similarity coefficient~\citep{niwattanakul2013using} between the generated texts and the training data.  Specifically, for each generated sentence, we compute its similarity with every sentence in the training data and select the highest one as the score. The average of the similarity scores is the overall novelty score. \textbf{BLEU}~\citep{BLEU} is a widely used metric for text generation, which calculates the n-gram overlap between generated texts and references. However, BLEU is inappropriate for open-ended generation task~\citep{fan_story_generation} in that there are multiple plausible texts for the same input, while only one is given as gold text. Here, we provide BLEU-4 scores for reference.

% and provides specific concepts
\medskip\noindent\textbf{Human evaluation}\hspace{1em}For each baseline, three annotators with linguistic background are required to score 100 samples randomly chosen from the test set. Each sample consists of a source sentence and two generated texts from our approach and a baseline. Annotators have no idea which system texts come from. They are asked to decide which output is better in terms of the following three aspects: \textbf{Fluency} measures whether the output is fluent, \textbf{Coherence} evaluates whether the output is semantically coherent, including topic sentence with generated sentences and generated sentences inside. \textbf{Informativeness} measures whether the output is diverse. The winner gets 2 points and loser gets 0 points. If two outputs are equally good or equally bad, they both get 1 points. 
We report the average score as the final result. 
% \section{Results and Discussion}
\subsection{Experiment Results}
% \medskip\noindent\textbf{Results}\hspace{1em} 
Table~\ref{table:automatic} presents automatic evaluation results.
% The results of the proposed approach and baselines are shown in Table \ref{table:automatic}. 
As expected, MRG outperforms  baselines by a large margin in terms of diversity and novelty. For instance, MRG-C achieves \textbf{22.6\%}, \textbf{55.8\%} relative improvements in terms of Dist-3
%and \textbf{0.3}, \textbf{0.5} absolute improvement in terms of Tri-J 
over the best baseline in story generation and review generation, respectively, with a small sacrifice in BLEU score. It demonstrates that with the help of  conceptual skeleton paths,
%inferred by multi-hop reasoning module,
MRG can generate more diversified and novel text than baselines while maintaining good readability .

Table~\ref{tab:human evaluation} demonstrates the results of human evaluation. Our approaches substantially outperform the baselines, especially in terms of informativeness and coherence. 
The conceptual skeletons extracted in our approaches help models conduct semantic transfer, making the generated text more coherent. Also, the inferred skeleton paths provide plenty of useful information which contributes to the informativeness of generated text.
%Our proposed model is able to extract conceptual skeleton paths via multi-hop reasoning on knowledge graphs to model sentence-level semantic dependencies, making the generated text more coherent. Besides, the inferred skeleton paths provide plenty of useful information which contributes to the informativeness of generated text.
%T-test results show that the geometric mean gaps are significant in most cases. 
MRG-S surpasses several strong baselines even without an artificial knowledge graph. It is interesting to see that MRG-S outperforms MRG-C in story generation. The main reason is that the story generation dataset is of higher-quality than the review  generation dataset and the Self-constructed graph contains much specific knowledge important for modeling semantic transfer.
\begin{figure}[t]
  \centering
  \small
  \footnotesize
  \includegraphics[width=1.0\linewidth]{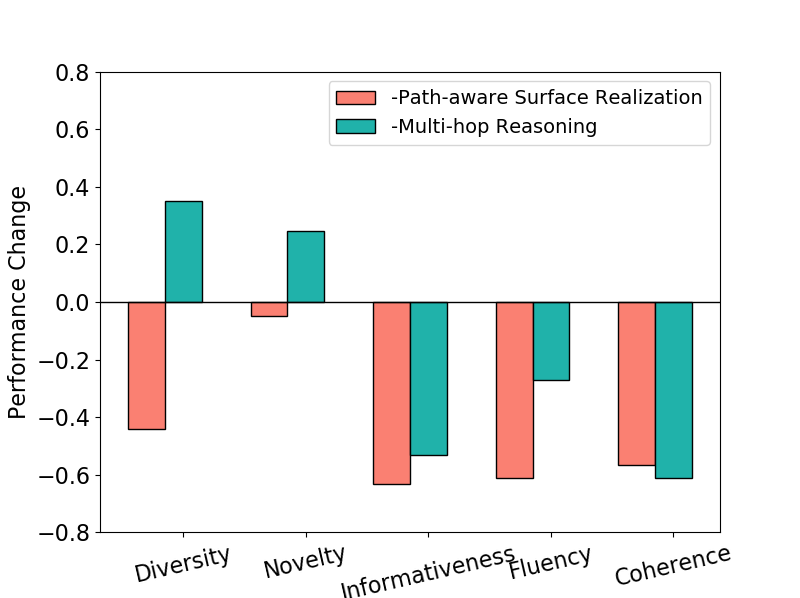}
  \caption{Ablation studies of MRG-C on story generation. Diversity is measured by Dist-3, novelty is measured by Tri-Jaccard. Informativeness, fluency and coherence is measured by human evaluation on 100 randomly sampled pairs. Y-axis shows the percentage of the performance change compared  with  MRG-C. 
  \label{fig:ablation}}
  \end{figure}

\subsection{Ablation Study}
Here we conduct  ablation studies to verify the effectiveness of two modules. Figure~\ref{fig:ablation}  demonstrates the results.

% \subsubsection{Multi-hop Reasoning Ablation}
\medskip\noindent\textbf{Multi-hop Reasoning Ablation}\hspace{1em}
Specifically, in reasoning stage we replace the  multi-hop reasoning module with a random reasoning module which randomly labels 1/3 concepts as target concepts, 1/3 as intermediate concepts. As we can see from Figure~\ref{fig:ablation}, the coherence and informativeness declines significantly when the reasoning module is removed. This illustrates that the multi-hop reasoning module is crucial to generating coherent text. The multi-hop reasoning module infers skeleton paths which bridge the  preceding sentence with the following ones, and random reasoning cannot model such semantic dependencies. 
% \subsubsection{Path-aware Sentence Realization Ablation}

\medskip\noindent\textbf{Path-aware Sentence Realization Ablation}\hspace{1em}
Specifically, we only use the target concepts to generate the subsequent sentence without using the conceptual path information. As shown in Figure~\ref{fig:ablation}, when the inferred paths are replaced with the target concepts, all five indicates drop, indicating that the intermediate path information are essential for improving the quality of generated text. The inferred conceptual skeleton paths provide plenty of information which is crucial to  informative and coherent text generation.

\subsection{Case Study}
\begin{figure}[t]
\centering
\small
\footnotesize
\includegraphics[width=0.9\linewidth]{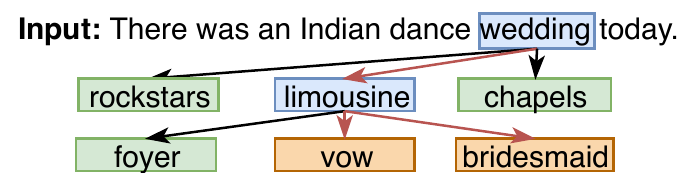}
% \includegraphics[\linewidth]{graph_trans.pdf}
% \vspace{-0.55in} 
\caption{An example of the inferred paths on the self-constructed graph. Blue words denote intermediate concepts. Orange words denotes target concepts. Green words denote other concepts.
\label{fig:reasoningpath}}
\end{figure}

Table~\ref{case} presents several examples  generated  by different systems. 
The baselines tend to generate low-quality texts in story generation. For example, texts generated by Seq2Seq and Transformer are very generic and suffer from bad inter-sentence coherence and repetition issues. Because they do not explicit model semantic dependencies among sentences. For GPT-2-FT, much of the content except for the first sentence is unrelated to the input. Besides, it suffer from repetition issues (e.g. \textit{went to beach for a day of fun}). Neither Skeleton2Seq nor GPT-2-KE can express information about keyword ``dance'' in the topic sentence. Although Plan\&Write can embody content about both keywords ``wedding'' and ``dance'', its output is relatively incoherent and less informative.
% e.g. ``the bride and groom were so happy to be married''. 
% By introducing external knowledge, Transformer-CC can generate more diversified text than vanilla Transformer. 
% However, due to without concept reasoning, the generated text lacks coherence and logic, e.g.,  ``The band played with the lead singer, the band played a trumpet player was very well.''. 
% For GPT-2, much of the content except for the first sentence is unrelated to the input. After enhanced with ConceptNet, GPT-2-KE yileds more specific topic words.
% the first few generated sentences is highly relevant with input, but as the text length grows, uncorrelated text appears, such as ``swimming'', ``play in the water'' is generated as the topic sentence is about ``Indian dance wedding'', showing long text generation is a challenging task, even the powerful pre-trained model is  discouraged in this field.  
In comparison, our methods can generate more diversified and coherent sentences with a multi-hop reasoning module to introduce external knowledge into sentence-level dependencies modeling.
% Besides, MRG-S generates more lower frequency words such as ``bridesmaids'' which doesn't exist in ConeptNet but in training data, showing how Self-constructed Graph help improve diversity on story generation. 
For a better demonstration, Figure \ref{fig:reasoningpath} presents the reasoning path, which provides an explanatory way to help understand how MRG works. 
As to product description, it can also be found that MRG reaches a better result than Seq2Seq and Transformer. Specifically, it has stronger ability in generating longer texts with abundant information. The case shows that MRG is able to provide a detailed description of the features.%, while the baselines demonstrate boring and repetitive expressions.

%%%%%%%%%%%%%%%%%%%%%%%%%%%%%%%%%%%%%%%%%%%%%%%%%\scaleboxconc
\begin{table}[!t]
\footnotesize
\small
\centering
\scalebox{0.92}{    \begin{tabular}{|p{1.00\linewidth}|}
    \hline

    \textbf{Input 2}: \textsl{there was an Indian dance wedding today.}\\
      \textbf{Seq2Seq}:  \textsl{the bride and groom were very happy to be married. the bride and groom danced the night away. the bride and groom danced the night away.}    \\  
        \textbf{Transformer}: \textsl{the bride and groom exchanged vows. the bride and groom were so happy to be married. the bride and groom were so happy to be married.}\\
      \textbf{Skeleton2Seq}: \textsl{the bride and groom are happy to be married. the bride and groom are happy to be married. the wedding party is very beautiful and happy. the little family pose for a photo.}\\ 
      \textbf{Plan\&Write}: \textsl{the groom was happy to be married. the bride and groom took their first dance. then they took a picture together. they danced the night away.}\\ 
      \textbf{GPT-2-FT}: \textsl{the bride and groom were very happy. the family went to the beach for a day of fun. they had a great time. they had a great time swimming. they had a great time playing in the water. the family went to the beach for a day of fun...[repetition]}\\ 
      \textbf{GPT-2-KE}: \textsl{the bride and groom were very excited. the ceremony was held indoors. the bride and groom had a great time. the flower girl was so happy.}\\

    \textbf{MRG}: \textsl{the bride dances with her bridesmaids. the bride and groom wore matching glasses. the bride and groom share a romantic kiss. the bride and groom exchanging vows. the bride and groom walked down the aisle.}\\ 
     \textbf{Ground Truth}: \textsl{there was a lot of dancing. the bride and groom 's first dance. it was all smiles and laughter. the whole family was there!}\\
    
    \hline%Dessert was the best part. Simple and tasty.

    % \textbf{Input 2}: \textsl{there was an indian dance wedding today.}\\

    \textbf{Input 2}: \textsl{2019 new Hepburn-style loose dress for summer, with lapels and beads, figure flattering.}\\
    %   \textbf{Seq2Seq}:  \textsl{the bride and groom were very happy to be married. the bride and groom danced the night away. the bride and groom danced the night away...[repetition]}    \\  
      \textbf{Seq2Seq}:  \textsl{This dress is designed with round collars , which helps show a pretty neck and highlights elegancy. The loose clothing pattern makes the dress comfortable to wear.}    \\
    %   \textbf{Transformer}: \textsl{the bride and groom exchanged vows . the bride and groom were so happy to be married. the bride and groom were so happy to be married...[repetition]}\\
      \textbf{Transformer}: \textsl{This dress is designed with round collars, which helps show a pretty neck and highlights elegancy. The classic round-collar design helps show a pretty neck and a pretty shape of face, and highlights elegancy.}\\
    %  \textbf{Transformer-K}: \textsl{the bride and groom were happy to be married. the bride and groom were so happy to be married. the bride and groom were so happy together.}\\
        %  \textbf{Transformer-CC}: \textsl{the bride and groom were very happy to be there. the bride and groom were very happy to be there. the bride and groom were very happy to be married. the bride and groom were very happy to be married.}\\
  %   \textbf{Template2Seq}: \textsl{The family was so happy to be there.}\\
    % \textbf{Skeleton2Seq}: \textsl{*}\\ 
    % \textbf{GPT-2}: \textsl{     the bride and groom were very happy. the family went to the beach for a day of fun. they had a great time. they had a great time swimming. they had a great time playing in the water. the family went to the beach for a day of fun...[repetition]}\\ 
    % \textbf{MRG-S}: \textsl{the bride dances with her bridesmaids. the bride and groom wore matching glasses. the bride and groom share a romantic kiss. the bride and groom exchanging vows. the bride and groom walked down the aisle.}\\ 
    \textbf{MRG}: \textsl{This dress is of a loose clothing pattern, which is suitable for most people. The design can provide you with more comforts and freedom, and the color demonstrates minimalist style. Dressing it up with a ripped pants makes you look tall and slim and demonstrates your elegancy. The fine workmanship ensures that the loose clothing pattern is casual and suitable for most figures.}\\ 
    %  \textbf{Ground Truth}: \textsl{there was a lot of dancing. the bride and groom 's first dance. it was all smiles and laughter. the whole family was there!}\\
     \textbf{Ground Truth}: \textsl{The dress with POLO collars makes you look outgoing in the photo. The beaded dress with a pair of silver pointe shoes is a perfect collocation for ladies. A simple black shoulder bag makes you look younger. The combination of sporty and debutante style is popular this year.}\\
    \hline
    \end{tabular}}
    \caption{Examples generated by the proposed approach and baselines. The top and bottom groups list the generated text on story generation and product description, respectively. The examples demonstrate that MRG can generate more coherent and informative text with less repetition in both tasks.}
    
    %The first one contains texts generated by different models on story generation, and the second one contains texts generated by models on product description generation.
   
    \label{case}
% 	\vspace{-0.10in}
\end{table}
%%%%%%%%%%%%%%%%%%%%%%%%%%%%%%%%%%%%%%%%%%%%%%%%%%%%%%

% %%%%%%%%%%%%%%%%%%%%%%%%%%%%%%%%%%%%%%%%%%%%%%%%
% \begin{table}[t]

% \footnotesize
% \centering
%     \scalebox{0.95}{\begin{tabular}{|p{0.95\linewidth}|}
%     \hline
%       \textbf{Input 1}: \textsl{every kind of musician was on stage.}\\
%     \textbf{MRG-C Reasoning path}: musician $\rightarrow$ band $\rightarrow$ concert.\\\hline
    
%     % \textbf{Input 2}: \textsl{The lincoln museum had a lot of information about his dead.}\\
%     % \textbf{Pivot concepts}: set \\\hline
    
%     \textbf{Input 2}: \textsl{there was an indian dance wedding today.}\\
%     \textbf{MRG-I Reasoning path}:   wedding $\rightarrow$ bridesmaid, wedding $\rightarrow$  rockstars $\rightarrow$ aisle.\\\hline
%     \end{tabular}}
%     \caption{The reasoning path.} 
%     \label{reasoningpath}
%     %\vspace{-0.1in}
% \end{table}
% %%%%%%%%%%%%%%%%%%%%%%%%%%%%%%%%%%%%%%%%%%%%%%%%%

\section{Related Work}
% In the following, we provide a review of the significant previous work in both multi-hop reasoning and text generation.  
\medskip\noindent\textbf{Multi-hop Reasoning.}\hspace{1em}Multi-hop reasoning has been drawing attention in the fields of natural language processing and machine learning.
The most representative tasks are Meta QA~\citep{zhang2018variational} and PathQuestion~\citep{zhou2018interpretable} in open-domain question answering over knowledge bases and Hotpot QA~\cite{yang-et-al:hotpotqa} in reading comprehension.
Recently, researchers have conducted a series of methods for multi-hop reasoning on graphs and obtained significant performance gains~\cite{zhou2018interpretable,qiu-et-al:multihop1,DBLP:conf/emnlp/JiangB19,DBLP:conf/acl/TuWHTHZ19,ding-et-al:multihop2}. Those methods formulate multi-hop reasoning as a sequential decision problem and solve it using GNNs~\citep{gcn,gat} or Memory Networks~\citep{memoryNetworks,DBLP:conf/icml/KumarIOIBGZPS16}. In this work, we propose to perform 
sequential decision on a knowledge graph with a powerful pre-trained model BERT. In addition, our work can be seen as the first step to study how multi-hop reasoning can help long text generation.
% However,  there still exists few studies about multi-hop reasoning for text generation, especially long text generation.   

\medskip\noindent\textbf{Long Text Generation.}\hspace{1em}The most popular models for text generation are Seq2Seq~\cite{sutskever-et-al:seq2seq} 
with attention~\cite{bahdanau-et-al:attention} 
and Transformer~\cite{vaswani-et-al:transformer}. 
However, these models have limited capacity in modeling complicated semantic dependencies among sentences~\cite{wiseman-etal-2017-challenges}, often suffer from logic conflicts.
%  and lack of long range coherence  in  generated text
There are two lines of work to tackle this challenge. 
The first line attempts to model long-term semantic dependencies among sentences by a huge-size model pre-trained on a large-scale dataset~\citep{radford-et-al:gpt2,unifiedlm,mass,gpt2-ke}. These methods can be categorized into implicit semantic dependencies modeling. 
% One is the implicit modeling semantic dependencies.~\citet{radford-et-al:gpt2,unifiedlm,mass} propose a huge-size model pretrained on data of large scale, which is able to model long-term semantic dependency. 
Another line is to explicitly model semantic relationship by simplifying sentence structures, i.e., models first plan a skeleton and then expand this skeleton to a long text. ~\citet{martin-event} extracted event sequences as skeletons.
% The second line of work approximates the semantic dependencies of the entire sentence by simplifying the sentence structure and then modeling only the semantic dependencies of the sentence skeleton. 
~\citet{xu-et-al:skeleton_jingjingxu_emnlp8} adopted a reinforcement learning method to extract sentence skeletons. ~\citet{li-et-al:coarse2fine} produced a syntactic sketch with additional part-of-speech tagging tasks. ~\citet{plan-and-write}  generated a sequence of keywords as planning conditioned upon the input.
 ~\citet{long-hunag} planed input items as a sequence of groups  and then realized each sentence conditioned on the planned result and the previously generated context. Different from those models, our methods build sentence-level dependencies by inferred skeleton paths on a knowledge graph, which is more interpretative and can cover more knowledge important for 
semantic transfer.
%  and can be see as the first attempt to introduce multi-hop reasoning into long text generation filed. 
%  Besides, our methods related to use knowledge graph to generate text~\citet{knowledge-one,knowledge-two,knowledge-three,knowledge-guan}.
% This paper explores the way to model semantic dependency among sentence via multi-hop reasoning in knowledge graph.

% \citet里面尽量不要写多个，如果同一作者还可以接受  : 好的
% Our work is also related to incorporating knowledge graph into text generation \cite{minliehung_commonsense_ijcai,graphTransformer,knowledgeablestoryteller}

\medskip\noindent\textbf{Knowledge-aware Text Graph}\hspace{1em}
% Our method is also related to use knowledge graph for text generation.
Recently, it has proven effective to integrate commonsense knowledge into text generation.
~\citet{knowledge-one} employed an additional knowledge encoder to improve dialogue system. ~\cite{knowledge-three} proposed to exploit commonsense knowledge for topic-to-essay generation with a memory-augmented generator.~\citet{knowledge-guan} adopted a multi-source attention mechanism to attend commonsense knowledge for story ending generation.~\citet{gpt2-ke} incorporated commonsense knowledge into pre-trained models during fine-tuning phrases. Despite their success, those  methods only attend triple-level 
knowledge and need the decoder to perform complicated  commonsense reasoning to leverage multiple triples for reasonable text generation. In comparison, our approach adopts a separate reasoning module to integrate multiple triples into a chain-like conceptual skeleton paths. The decoder just needs to focus on  generating a fluent sentence based on the inferred paths, which is more effective.
% Different from these models, our approach infers a skeleton path from knowledge graph via a reasoning module, which 
% to help long text generation.

\section{Conclusions}
% In this paper, 
% we study how multi-hop reasoning improves long text generation and propose a 
In this paper, we propose a
novel multi-hop reasoning generation model to generate coherent and informative long text via modeling complicated semantic dependencies among sentences. We adopt a self-supervised way to construct the supervisory signals to train the new model.
% This work can be viewed as the first step towards using multi-hop reasoning for long text generation.
% This work can be viewed as the first step towards multi-hop reasoning for long text generation. 
We evaluate our model on three representative long text generation tasks, and propose to build a Self-constructed graph to make the new model applicable on tasks without appropriate artificial knowledge graphs. Automatic and human evaluation show that the new model achieves substantial improvements  over a series of state-of-the-art baselines, such as pre-trained models and knowledge-enhanced models, in terms of coherence, informativeness, and novelty. 
% Besides, our methods can generete entitiy diversity text and the reasoning path that provides a explanatory view for long text generation. 
% Besides, the analyses demonstrate that our approach can generate texts of higher diversity and novelty. 
Furthermore, the inferred paths enhance the interpretability of text generation. In the future, we will elaborate the reasoning mechanism to support controllable text generation. Besides, we will make the two modules interact with and promote each other to  avoid
error propagation. 
% in the current pipeline generation system 
% In the future work, we attempt to elaborate more reasoning mechanism. 
\bibliography{aaai21}

\clearpage
\appendix

\end{document}